# Using Computer Vision to enhance Safety of Workforce in Manufacturing in a Post COVID World

*Prateek Khandelwal[1], *Anuj Khandelwal[1], *Snigdha Agarwal[1], Deep Thomas[1], Naveen Xavier[1], Arun Raghuraman[1]

[1] Group Data & Analytics, ABMCPL, Aditya Birla Group, India
*Equal Contribution

*Abstract* - The COVID-19 pandemic forced governments across the world to impose lockdowns to prevent virus transmissions. This resulted in the shutdown of all economic activity and accordingly the production at manufacturing plants across most sectors was halted. While there is an urgency to resume production, there is an even greater need to ensure the safety of the workforce at the plant site. Reports indicate that maintaining social distancing and wearing face masks while at work clearly reduces the risk of transmission. We decided to use computer vision on CCTV feeds to monitor worker activity and detect violations which trigger real time voice alerts on the shop floor. This paper describes an efficient and economic approach of using AI to create a safe environment in a manufacturing setup. We demonstrate our approach to build a robust social distancing measurement algorithm using a mix of modern-day deep learning and classic projective geometry techniques. We have deployed our solution at manufacturing plants across the Aditya Birla Group (ABG). We have also described our face mask detection approach which provides a high accuracy across a range of customized masks.

*Index Terms*- computer vision, deep learning, manufacturing, safety, COVID-19, social distancing, face mask detection, neural networks

## I. INTRODUCTION

The spread of COVID-19 [1] virus and the ensuing largescale lockdowns across the globe has given rise to an alarming situation. The resumption of production in manufacturing setups across all sectors is a key pre-requisite for kickstarting economic activity of a nation. While there is an urgent need to resume operations at these plants, the safety of the workforce operating these plants cannot be compromised. Accordingly, processes are being put in place to educate the workforce regarding new safety regulations at the workplace which helps reduce the risk of virus transmission. However, to help the workforce transition into a post COVID world, there was a need for us to build solutions that help monitor and alert individuals once a safety violation occurs.

All plants have CCTV installation, with at least a few hundred cameras as part of their security system setup. It is however not practical to monitor all these feeds concurrently due to the manual nature of the task. We have built a system that takes in these feeds and analyzes frames using deep learning models to detect whether violations have occurred or not. Once detected, a real time voice alert is triggered in the area of the violation. This feedback helps reduce the violations and thus contributes to the overall safety at the plant. In addition, these alerts are stored in a central repository that helps the management analyze the trends and take suitable actions to curb the violations. An overview of the solution is provided in Section III of this paper.

Given the context of COVID-19, we focused on building features that help reduce the risk of virus transmission. Research [2] indicated that maintaining social distance between co-workers as well as wearing face masks were effective means of reducing this risk. We hence built solutions that could monitor these actions through video feeds.

World Health Organization (WHO) has recommended that a social distance of at least 2m [3] be maintained between individuals. While the requirement is simple, monitoring this aspect through video feeds that provide a perspective view makes it difficult to ascertain the exact distance on ground. We provide the details of the model and its aspects in Section IV of the paper.

WHO has also recommended that personnel are encouraged to wear face masks to avoid the risk of virus entering the body through the nasal / oral cavity [3,4]. During the lockdown, it was encouraged by the Indian government for the people to come up with mask substitutes [4] as most countries saw a scarcity of required PPE [5]. Hence, the face masks worn are not of standard type and come in different colors, shapes and sizes. The lack of such diversified data for training purposes makes mask detection a challenging task. The approach taken to overcome this problem, the model and other details are discussed in Section V of the paper.

## II. RELATED WORK

While there has been a lot of discussion around deep learning-based approaches for person detection [23], we did not find any work on distance measurement between these detected persons. This encouraged us to come up with our own algorithm to solve this problem. Our work on face mask detection comprises of data collection to tackle the variance in kinds of face masks worn by the workers. The face mask detection model is a combination of face detection model to identify the existing faces from camera feeds and then running those faces through a mask detection model. We summarize below the existing and related work for both the models.

### A. PERSON DETECTION MODELS

For all our models, we have used MobileNetV2 as the core model for detection. This is because we require a framework with models to carry out processing at a minimum speed of 2-3 frames per second. MobileNetV2 provides a huge advantage in computation cost as compared to a normal 2D convolution model. A deeper dive into the comparison results has been established in [15, 19]. We go by the results shared by the paper in considering the MobileNetV2 model.

### B. FACE DETECTION MODELS

The Viola Jones Face Detector [7] which used cascaded Haar features is one of the most widely used face detector. Another model for face detection was proposed by Li et al. [8] which was a Multi-View Face Detector using surf features. In addition, a tree structure based deformable part model was proposed in [9]. However, with the advent of deep learning, CNN based face detectors [6] are being widely used by the research community and industries alike. CNN based models learn face representations from the annotated data as described in [10]. These models are then capable of identifying faces and suggesting facial landmarks in the input images.

### C. MASK DATASETS

With the increase in the importance of wearing face masks during the COVID-19 pandemic, there are a plethora of blogs and social media posts that describe approaches to the mask detection problem. However, most of these approaches do not provide access to a reasonably large dataset that works for real world use cases. Most of the work available uses smaller datasets with less variability in terms of types of masks (i.e. standard N-95, surgical masks) or only look at specific regional datasets that have been scraped from the web.

### D. MASK DETECTION MODELS

Our work comes very close to the work described in [11]. The cited work looks at identifying people with full face or partial occlusion. The paper categorizes people with hand over their faces or occluded with objects. This approach is unsuitable for our scenario which requires to essentially detect faces that have their mouths covered with mask like objects such as scarves, mufflers, handkerchiefs etc.

## III. SOLUTION OVERVIEW

This section briefly describes the solution architecture and how the system functions in an integrated manner to achieve the objective of ensuring safety of workers at our manufacturing plants. There are many ways to solve the problem of ensuring safety of workers in the workplace which include physical monitoring as well as leveraging wireless technologies such as Bluetooth, RFID etc. However, we chose to leverage existing CCTV infrastructure at the plant for the following reasons:

1. Provides adequate coverage of the premises.
2. Non-intrusive for the workforce.
3. Minimal hardware requirements at the plants.

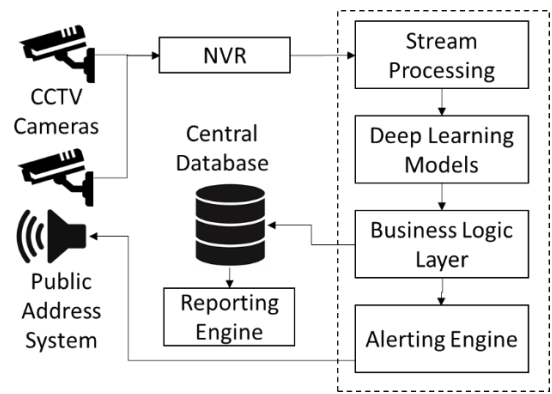

**Figure 1**: *Solution architecture diagram which shows the main components that have been integrated and deployed at our plants.*

The CCTV camera video feeds from the Network Video Recorder (NVR) are streamed to a central server using RTSP. The concurrent streams are processed using *ffmpeg* libraries and queued across individual deep learning inferencing models. The output of each model is processed through a business logic layer and relevant alerts are sent to the alerting engine. The alerting engine (AE) orchestrates the voice alerts that need to be played at a specific site with a specific message. The alerts are also stored in a central repository that can be accessed by the management to carry out further analysis.

## IV. SOCIAL DISTANCING MONITORING

The social distancing monitoring module resides as part of the model layer in the solution. In case any video feed records any recurring violations lower than the threshold distance value for minimum specified duration, an alert is triggered to the AE. The alert is subsequently handled by the AE to trigger the necessary mechanism. This can also enable audit traceability by linking all associations between infected individuals. This module detects workers in a given frame and then calculates the inter-se distance between each worker. The details of each section are discussed in this section.

### A. Data Collection and Processing

We collected live videos feeds from CCTVs cameras across various manufacturing sites of the Aditya Birla Group and generated a large dataset of frames that was used for training our models. Visual inspection of the videos showed that our target object, viz. a person, was not present in all the frames. In order to avoid going through all feeds and manually selecting frames containing persons, we filtered videos to select frames that have moving objects which majorly constitutes working staff. We make use of the existing background subtraction algorithm as described in [24, 27]. As a pre-processing step, we also categorized frames containing objects in low-light and normal light. Finally, a total of 6589 person annotations were completed of which 80% were set aside for training and the remaining 20% for validation.

*B. Person Detection*

We have made use of the transfer learning approach [20] and have fine-tuned MobileNetV2 [19] model for person detection using TensorFlow framework [21]. We finally convert the trained model to Intermediate Representation (IR) format for OpenVINO Deep Learning Workbench [22]. As the MobileNetV2 paper confirms that it utilizes standard operations present in all neural frameworks, it was easy for us to convert native TensorFlow from binary format to the OpenVINO (IR) format.

We trained the model using TensorFlow Object Detection API. We used $300 \times 300$ size input for our SSD model with MobileNetV2 as backbone. We trained the model for 500,000 steps with *rmsprop* optimizer. The parameters were configured to $lr = 0.004$ with decay factor $\rho = 0.95$. After training, the final results achieved were $mAP@.50\ IOU = 0.897$ and $mAP@0.75\ IOU = 0.505$.

*C. Distance Measurement*

We run our custom trained MobileNetV2 model for person detection on each frame. The output from the model is a list of coordinates of bounding rectangles on detected persons; where a single rectangle is represented as: *[xmin, ymin, width, height]*. To find out if the detected people are following social distancing, we need to calculate the Euclidean distance between the bounding rectangles. The imaging plane or the camera plane is a 2D projection of 3D world coordinates therefore, the spatial relationship between the objects in this plane changes due to camera perspective. The objects near to the camera appear larger than those that are away from it. Calculating the distance between the rectangles in this perspective would give an incorrect estimate of the actual distance, we must correct it by transforming the image into top-down-view or bird's-eye-view. This is done by applying an Inverse Perspective Transform onto the image. In essence, we compute the Homography Matrix (denoted here as M) which is used to carry out the transformation. There are three parts to the calculation of M, and are detailed below [26]:

1. Translation (to make image zero-centered), If $x, y$ be coordinates in image plane, the shifted image coordinates will be
$$X = x - \frac{IMAGE_{WIDTH}}{2}$$
$$Y = y - \frac{IMAGE_{HEIGHT}}{2}$$

2. Rotation and Scaling is computed along 2 planes ($Rx \times Rz$)

$$R = \begin{bmatrix} S_x(cos\emptyset) & 0 & -sin\emptyset \\ 0 & S_y & 0 \\ sin\emptyset & 0 & cos\emptyset \end{bmatrix}$$

where $\emptyset$ is the angle between camera plane and perspective view plane; $S_x$ and $S_y$ are scaling factors to resize stretched image back to its original dimension.

3. Projection is last step to convert this 3D transformed image to 2D form assuming $(p, q, r)$ are 3D coordinates in a rotated image and $(u, v)$ in converted 2D image and assuming $f$ is focal length of camera

$$u = \frac{fp}{f-r} + \left(\frac{IMAGE_{WIDTH}}{2}\right)$$
$$v = \frac{fq}{f-r} + \left(\frac{IMAGE_{HEIGHT}}{2}\right)$$

In this work, we have used OpenCV implementation to compute the transformation matrix, M. OpenCV mapping is done from destination to source and thereafter inverse of the matrix is taken [28].
$$\tilde{x} = Mx$$
$$x = M^{-1}\tilde{x}$$
where $\check{x}$ are the coordinates when seen through a bird's eye view.

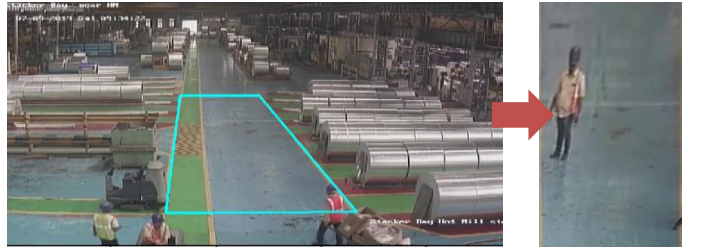

**Figure 2**: Perspective transformation of selected area

We compute matrix M by selecting four points in our image which is in perspective or camera view, such that none of the three points make a co-linear set. This forms our region of interest bounded by a four-edged irregular quadrilateral. The selection of these points is done in such a way that for each of edges we know its actual length on the ground. Here, we make the assumption that the region of interest chosen in the image is large enough to be able to approximate the projection of all the points onto a top-view image. This entails that we can use the transformation matrix M to obtain the projection of any given (x, y) coordinate in this image plane into our desired projection plane, which in this case is the top-view image.

Using the matrix M, we project all the bounding rectangles obtained from person detection algorithm. Since the bounding rectangles are of the people standing upright, it is safe to assume that the bottom coordinates of a rectangle, would naturally correspond to the ground points for the detected person. Hence, we pick the bottom coordinates of the rectangles and compute Euclidean distance between them. For a pair of rectangles each having two ground points, we calculate four linear distances

Euclidean distance $E(X,Y) = \sqrt{(x1-x2)^2 + (y1-y2)^2}$
$d(P,Q) = min\big(E(p1,q1), E(p1,q2), E(p2,q1), E(p2,q2)\big)$
where $P = [p1, p2]$ and $Q = [q1, q2]$ are set of ground points. From each individual $dij$ distance matrix $D$ is computed. $D$ is a zero-diagonal symmetric matrix

$$D = \begin{bmatrix} d_{11} & \ldots & d_{1n} \\ \vdots & d_{pq} & \vdots \\ d_{n1} & \ldots & d_{nn} \end{bmatrix}$$

Thereafter the minimum of the four distances is chosen and compared to threshold distance "*t*" which is the assumed safe

ecided

distance. If the minimum calculated distance is less than the threshold, we generate an alert thus ensuring that no two sets of people stand close to each other.

### D. Results

To test the performance of our transformation matrix, we took a sample image in perspective view and measured the dimension manually on ground.

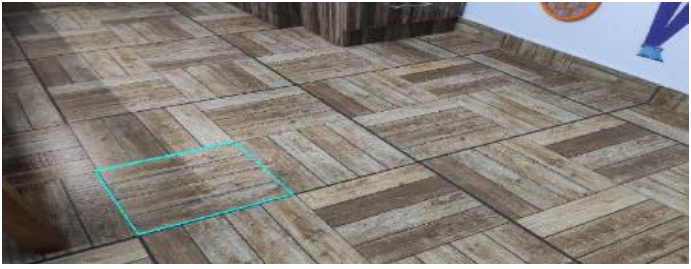

**Figure 3:** *Calibrating and measuring ground points to create transformation matrix*

In the picture above, the actual dimensions of the tile are 30 cm x 30 cm square. With input and output coordinated data available, we calculated transformation matrix $M$, described above. To measure goodness of our method we use transformation matrix and calculated distance between two points across X, Y and X-Y direction.

| Distance from square ($units \times 30cm$) | 1 | 2 | 3 | 4 |
|---|---|---|---|---|
| X | 30.32 | 30.41 | 30.19 | 30.53 |
| Y | 29.33 | 28.51 | 28.05 | 27.5 |
| X-Y | 29.26 | 28.72 | 27.39 | 27.05 |

**Table 1** *: Measurement comparison across X,Y and X-Y direction*

In the experiment described above, we chose two points using a mouse, which would inadvertently introduce human error during points selection. Despite the unavoidable error, we can see from **Table *1*** that the results have more than 90% accuracy. As the size of reference quadrilateral increases, transformation matrix M, becomes more generalized, and the overall testing error decreases.

As described in Section II, OpenVINO provides synchronous and asynchronous mode which enable faster inferencing. We profiled the computation for each detection in both modes and grouped it with respect to number of people detected. We have plotted the boxplot to show gain in asynchronous mode when compared with synchronous mode.

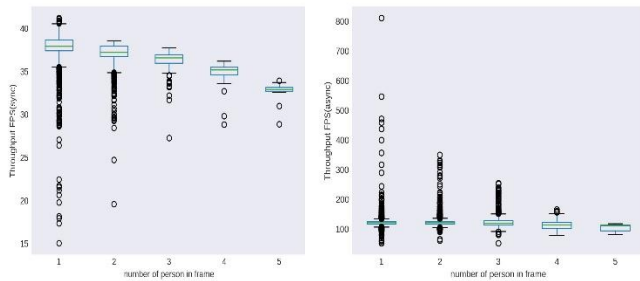

**Figure 4:** *FPS comparison across Sync and Async mode*

To compare both modes in the same scale, we took median of the data for number of persons detected. We can see in Table 2 that asynchronous mode is at least 3 times faster than synchronous mode.

| # of persons(detected) | 1 | 2 | 3 | 4 | 5 |
|---|---|---|---|---|---|
| *Median FPS(sync)* | 38 | 37 | 37 | 35 | 33 |
| *Median FPS(async)* | 122 | 122 | 121 | 115 | 112 |

**Table 2**: *FPS comparison between Sync and Async mode*

Bearing in mind the optimization and calibration methods described above, we deployed the algorithm with thresholds so that it triggers an alarm once social distancing violation occurs.

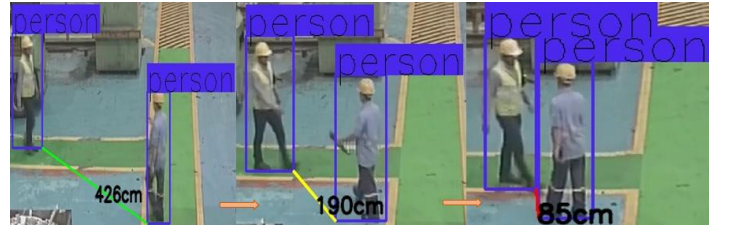

**Figure 5:** *Calculation of distances between two workers in real time and detection of violation once calculated distance is below threshold distance*

### E. Limitations

Distance between persons can be calculated only if all persons are detected clearly. In case of overlap, person detection will fail and the distance cannot be calculated. Accuracy of the distance is based on transformation matrix, which in turn depends on accurate calibration of ground points. It has been observed that slight human errors during calibration will result in large errors during distance computation.

## V. FACE MASK DETECTION

The face mask monitoring module also resides as part of the model layer in the solution. Once a worker is detected who is not wearing a face mask, an alert is again triggered to the AE. This module accepts individual frames and are treated as static images. The processing of these images is carried out in two steps; Face Detection and Mask Classification. The approach taken to build these models is described in this section.

### A. Face Detection

We used deep learning-based CNN models for face detection due to the advantages have been established widely and described in [12, 13, 14]. We used an off-the-shelf pre-trained MobileNetV2 model and observed a model accuracy of 94.1% with a head-height of greater than 100 pixels. The minimum face size that can be detected by this model is 90 X 90 pixels. The output of the model is used to crop-out the face bounding box with an offset of 30 pixels from the model output. This face crop is then used as an input to the Mask Detection model.

### B. Data Collection

For training the mask detection model, we used the Face Detection model as described above to extract face-crops from few video feeds of a single manufacturing plant. These face-crops of about 500 images were manually annotated into *mask* and *no mask* categories. We built a rudimentary deep learning off-the-shelf CNN classifier on this dataset. We then ran the face detection model on the video feeds from multiple plants and from cameras ranging at different heights. We then used our rudimentary model as a semi-supervised [16] training approach to now label the remaining cropped faces. This is also in close accordance with the handling of partially labelled training data as described in [17]. Our dataset consists of a total of 4225 annotated images with 1900 in class *mask* and 2300 in class *no mask*. The sample data collected is shown in the figure below.

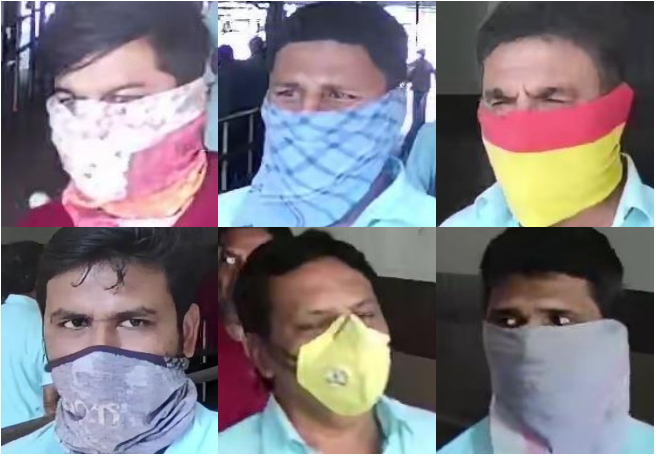

*Figure 6: Sample data containing diverse face masks*

### C. Mask Detection

We used the above annotated images to train a deep learning binary classification model as described in [18] that classifies an input image into *mask* and *no mask* categories using the output probability of the two classes. We used 20% of our overall data as a validation set with 380 as *mask* and 460 images as *no mask* which is not exposed during training of the model.

We used the MobileNetV2 architecture for building the model. We resized the images to $224 \times 224$ pixels to be fed into the network. We have used an augmentation strategy to bring variation in the data to handle the images from the cameras ranging from 5 feet to 10 feet. We have restricted to minor augmentations in terms of horizontal flip, zoom factor, shear, image brightness, height and width shift. These augmentations help us in building a robust model that can handle varying lighting and wall installation conditions of the cameras at different plants. We use the Adam optimizer with initial β-parameters of $\beta_1 = 0.0009$, $\beta_2 = 0.999$ and learning rate $1 \times 10^{-4}$ which is modified at the plateau. The parameters were identified using a fine search strategy between initial values obtained using a hit and trial methodology. We sampled batches using a fixed batch size of 8 images. We trained the network for 40 epochs, saving checkpoints every 200 iterations and dropping learning rate by a factor of 0.1 at plateaus with a patience period of 5 epochs.

### D. Results

The AUROC [25] of the model is 97.6%. The precision is 0.97 and the recall is 0.97 with *mask* being the positive class. The frames per second throughput of the model is 42.09.

### E. Limitations

We observed a few limitations in our current model. The model is not able to correctly classify face images where the face is partially hidden by a person in the front as shown in Figure 7. Also the model is not able to detect faces if the camera height is greater than 10 feet.

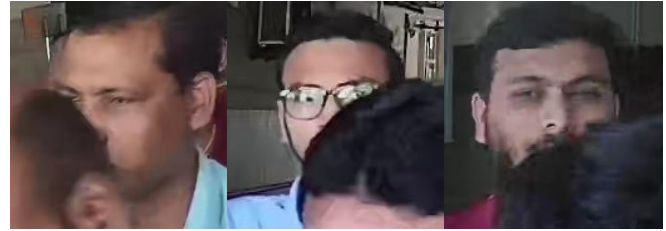

**Figure 7:** *Shows faces which get incorrectly identified as the face gets occluded by the person in the front*

## VI. FUTURE WORK

The use cases discussed above are only some of the many features that have been integrated as part of this solution. We believe that there are a host of other use cases that can be included in this suite to provide a more holistic sense of safety at our plants. Some of the features, currently under development are briefly discussed below:

*A. Temperature Detection*

High temperature due to fever has been identified as one of the symptoms of a corona virus infection. Identifying workers who have fever at the entrance and quarantining the individual is an essential practice that needs to be followed. Using handheld IR sensors to measure the temperature of each worker is a time-consuming process. Given the large influx of workers during a shift change this manual procedure leads to loss of productive time.

We are now working on integrating cameras that provide dual feeds; viz the RGB feed from the CMOS sensor and the thermal feed from the VOX (Vanadium Oxide) sensor. We can carry out face detection using the RGB feed and determine the temperature from the ROI of the forehead that is visible above the mask. Thermal cameras are available that provide a temperature reading with an accuracy of +/- 0.5°C at a distance of 2m from the camera. Once a temperature reading beyond the defined threshold is detected, an alert in the form of an audio signal is triggered which enables security personnel at the entrance to take suitable measures.

*B. Contactless Attendance*

Workers entering the factory premises at the entrance are required to mark their attendance using a fingerprint biometric scanner. The use of such systems poses a high risk in a post COVID world as the scanner surface becomes a potential medium for spread of the virus.

In view of the above, we are proposing our plants to migrate to contactless attendance using facial recognition technology. We are training deep learning models to detect faces and extract embeddings which are then matched against a repository of embeddings created using the HR database. These efforts are showing promising results and are likely to be deployed in production soon.

## VII. CONCLUSION

In this paper, we propose an approach that uses computer vision to monitor the activity of workers using CCTV feeds and ensures safety of the workforce in a manufacturing setup. We also discussed in detail the social distancing monitoring and face mask detection that helps maintain a safe environment in our plants in the post COVID world.

The deployment of this solution has been successfully tested and operational at some of our plants at Aditya Birla Group. The solution has the potential to reduce violations significantly through real time interventions. We believe that this approach will not only increase the safety at plants but also enhance the efficiency of plant processes in the time to come.

### ACKNOWLEDGMENT

We would like to take this opportunity to thank Group Data & Analytics for giving us the opportunity to work on this project. We also would like to acknowledge the support of Shrijeet Mishra, Innovation Head of the Aditya Birla Group without whom we would not have been able to take this forward.

### REFERENCES

[1] Wang Chen, Horby Peter W, Hayden Frederick G, Gao George F. A novel coronavirus outbreak of global health concern. The Lancet. 2020;395(10223):470–473. doi: 10.1016/S0140-6736(20)30185-9.

[2] Matrajt L, Leung T. Evaluating the effectiveness of social distancing interventions to delay or flatten the epidemic curve of coronavirus disease. Emerg Infect Dis. 2020

[3] https://www.who.int/emergencies/diseases/novel-coronavirus-2019/advice-for-public

[4] https://www.cdc.gov/coronavirus/2019-ncov/prevent-getting-sick/cloth-face-cover.html

[5] https://www.who.int/news-room/detail/03-03-2020-shortage-of-personal-protective-equipment-endangering-health-workers-worldwide

[6] Xu, L. & Ren, Jimmy & Liu, C. & Jia, J.. (2014). Deep convolutional neural network for image deconvolution. Advances in Neural Information Processing Systems. 2. 1790-1798.

[7] P. Viola and M. Jones, "Rapid object detection using a boosted cascade of simple features," Proceedings of the 2001 IEEE Computer Society Conference on Computer Vision and Pattern Recognition. CVPR 2001, Kauai, HI, USA, 2001, pp. I-I.

[8] J. Li and Y. Zhang. Learning SURF cascade for fast and accurate object detection. In IEEE CVPR, pages 3468–3475, 2013.

[9] X. Zhu and D. Ramanan. Face detection, pose estimation, and landmark localization in the wild. In IEEE CVPR, pages 2879–2886, 2012.

[10] S. S. Farfade, M. J. Saberian, and L. Li. Multi-view face detection using deep convolutional neural networks. In ACM ICMR, pages 643–650, 2015

[11] S. Ge, J. Li, Q. Ye and Z. Luo, "Detecting Masked Faces in the Wild with LLE-CNNs," 2017 IEEE Conference on Computer Vision and Pattern Recognition (CVPR), Honolulu, HI, 2017, pp. 426-434.

[12] GOODFELLOW, I., BENGIO, Y., & COURVILLE, A. (2016). Deep learning. Chapter 6.

[13] Winkler, David. (2018). D`10.1002/9783527816880.ch11_03.

[14] The History Began from AlexNet: A Comprehensive Survey on Deep Learning Approaches, Md Zahangir Alom, Tarek M. Taha, Chris Yakopcic, Stefan Westberg, 2018

[15] Howard, Andrew & Zhu, Menglong & Chen, Bo & Kalenichenko, Dmitry & Wang, Weijun & Weyand, Tobias & Andreetto, Marco & Adam, Hartwig. (2017). MobileNets: Efficient Convolutional Neural Networks for Mobile Vision Applications.

[16] Zhu, X. 2006. Semi-supervised learning literature survey. Computer Science, University of Wisconsin-Madison 2(3):4.

[17] Wu, B.; Lyu, S.; Hu, B.-G.; and Ji, Q. 2015. Multilabel learning with missing labels for image annotation and facial action unit recognition. Pattern Recognition 48(7):2279–2289

[18] Bellinger, Colin & Sharma, Shiven & Japkowicz, Nathalie. (2012). One-Class versus Binary Classification: Which and When?. Proceedings - 2012 11th International Conference on Machine Learning and Applications, ICMLA 2012. 2. 102-106. 10.1109/ICMLA.2012.212.

[19] M. Sandler, A. Howard, M. Zhu, A. Zhmoginov and L. Chen, "MobileNetV2: Inverted Residuals and Linear Bottlenecks," 2018 IEEE/CVF Conference on Computer Vision and Pattern Recognition, Salt Lake City, UT, 2018, pp. 4510-4520.

[20] S. J. Pan and Q. Yang, "A Survey on Transfer Learning," in IEEE Transactions on Knowledge and Data Engineering, vol. 22, no. 10, pp. 1345-1359, Oct. 2010.

[21] Martín Abadi, Ashish Agarwal, Paul Barham, Eugene Brevdo, Zhifeng Chen, Craig Citro, Greg S. Corrado, Andy Davis, Jeffrey Dean, Matthieu Devin, Sanjay Ghemawat, Ian Goodfellow, Andrew Harp, Geoffrey Irving, Michael Isard, Rafal Jozefowicz, Yangqing Jia, Lukasz Kaiser, Manjunath Kudlur, Josh Levenberg, Dan Mané, Mike Schuster, Rajat Monga, Sherry Moore, Derek Murray, Chris Olah, Jonathon Shlens, Benoit Steiner, Ilya Sutskever, Kunal Talwar, Paul Tucker, Vincent Vanhoucke, Vijay Vasudevan, Fernanda Viégas, Oriol Vinyals, Pete Warden, Martin Wattenberg, Martin Wicke, Yuan Yu, and Xiaoqiang Zheng. TensorFlow: Large-scale machine learning on heterogeneous systems, 2015. Software available from tensorflow.org.

[22] A. Demidovskij et al., "OpenVINO Deep Learning Workbench: Comprehensive Analysis and Tuning of Neural Networks Inference," 2019 IEEE/CVF International Conference on Computer Vision Workshop (ICCVW), Seoul, Korea (South), 2019, pp. 783-787.

[23] Z. Zhao, P. Zheng, S. Xu and X. Wu, "Object Detection With Deep Learning: A Review," in IEEE Transactions on Neural Networks and Learning Systems, vol. 30, no. 11, pp. 3212-3232, Nov. 2019, doi: 10.1109/TNNLS.2018.2876865.

[24] S. S. Mohamed, N. M. Tahir and R. Adnan, "Background modelling and background subtraction performance for object detection," 2010 6th International Colloquium on Signal Processing & its Applications, Mallaca City, 2010, pp. 1-6, doi: 10.1109/CSPA.2010.5545291.

[25] Andrew P. Bradley. 1997. The use of the area under the ROC curve in the evaluation of machine learning algorithms. Pattern Recogn. 30, 7 (July, 1997), 1145–1159. DOI:https://doi.org/10.1016/S0031-3203(96)00142-24

[26] M.Venkatesh, P.Vijayakumar. Transformation Technique. International Journal of Scientific & Engineering Research, Volume 3, Issue 5, May-2012

[27] M. Piccardi, "Background subtraction techniques: a review," *2004 IEEE International Conference on Systems, Man and Cybernetics (IEEE Cat. No.04CH37583)*, The Hague, 2004, pp. 3099-3104 vol.4, doi: 10.1109/ICSMC.2004.1400815.

[28] https://docs.opencv.org/2.4/modules/imgproc/doc/geometric_transformations.html

## AUTHORS

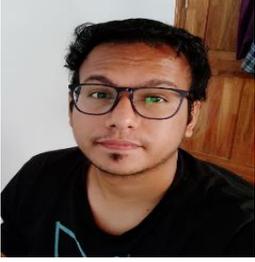

[Prateek Khandelwal](#) is an Engineering Design graduate with a minor in Industrial Engineering from IIT, Madras. He is currently working for Group Data & Analytics, Aditya Birla Group as an Artificial intelligence engineer. He has hands-on experience of over 5 years in solving analytics problems related to predictive modelling, statistical modeling, natural language processing and computer vision in the domain of manufacturing, smart-cities and retail industry. In the past he has also worked with the Bosch Center for Artificial intelligence.

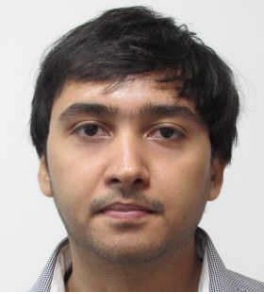

[Anuj Khandelwal](#) is Computer Science Engineer with a rich experience of solving problems at the intersection of Computer Vision and Machine Learning. In his current role, he works as a Data Scientist with Group Data & Analytics, Aditya Birla Group. In the past, Anuj has worked with Quantela Inc., a company focused on building an AI platform for Smart Cities and with Mu Sigma Inc., where he was one of the core members of the R&D lab and delved into creating cutting edge solutions in the area of Computer Vision, Internet of Things (IoT), Machine Learning and Big Data. Anuj was also a part of elite team at IIIT Delhi as a Research Assistant where he worked on Advanced driver-assistance systems(ADAS).

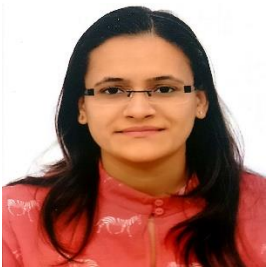

[Snigdha Agarwal](#) is a post-graduate in Data Science from VIT University, Vellore with a total experience of 5+ years working as an AI/ML engineer. She has worked with the R&D departments of Wipro Technologies (Digital) for banking, retail clients and Siemens Healthineers (Healthcare) for developing cutting edge Deep Learning and Machine Learning algorithms for imaging modalities using computer vision. In her current role with Aditya Birla, she is working with computer vision department for use-cases in Video Analytics. She is currently pursuing her Ph.D from IIIT Bangalore.

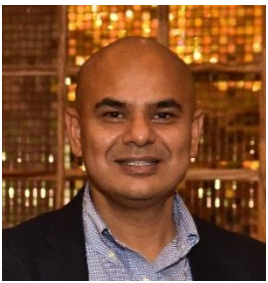

[Deep Thomas](#) is a reputed analytics expert, thought leader and a passionate evangelist of data science with a distinguished career spanning more than two decades of delivering sustained and increasing profitability through analytics-led business transformation. As Chief Data & Analytics Officer at Aditya Birla Group; Deep is leading the charge of its AI-led digital transformation to enable operational efficiencies, drive new business growth, and deliver successful outcomes across its multi-sector business. Prior to Aditya Birla Group, Deep was the Founder-CEO of Tata Insights & Quants, Tata Group's Big Data and Decision Science company; thereby giving him a unique privilege of setting up the Analytics arm for the founder institutions of modern India, Tatas and Birlas. Deep has also held various key positions in US and India with multinationals like Citigroup, HSBC, and American Express to steer their Global Digital and Analytics agenda. He is a Computer Science graduate with an MS in Information Systems and an MBA in Marketing from Arizona State and holds patents in Decision Science and Information Management capabilities.

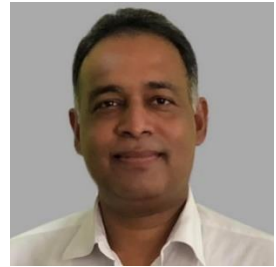

[Naveen Xavier](#) is a veteran from the Indian Army with over two decades of experience across a wide spectrum of technologies. He was part of the core technology team at NATGRID under the Ministry of Home Affairs. He also co-founded dataval Analytics, a niche AI/ML services consulting company. He now heads the Data & Analytics Platforms vertical at the Aditya Birla Group where he is responsible for creating AI/ML products to enhance customer experience, improve process efficiency and reduce cost. He has a keen interest in applying AI techniques in the cognitive domain and evangelizes its adoption across the enterprise.

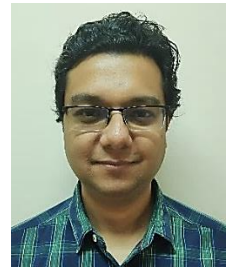

[Arun Raghuraman](#) is an analytics professional and a product enthusiast with a rich experience of 11+ years in Software and Analytics. He has experience in development, integration and solution design in Business consulting to provide analytics based solutions for various business problems and in his current role is responsible for managing the analytics products & platforms for Group Data and Analytics within Aditya Birla Group. In his previous roles he has been an analytics solution architect for Accenture, designing multiple solutions across various industries to drive productivity and profitability through analytics. Prior to that he has worked on the productization and implementation of analytics projects to drive business values with clients and enable data driven decision making strategies.